\pdfoutput=1

\documentclass[11pt]{article}

\usepackage[]{EACL2023}

\usepackage{times}
\usepackage{latexsym}

\usepackage[T1]{fontenc}

\usepackage[utf8]{inputenc}

\usepackage{microtype}

\usepackage{inconsolata}

\usepackage{amsmath}
\usepackage{amssymb}
\usepackage{booktabs}

\usepackage{pbox}
\usepackage{makecell}
\usepackage{multirow}
\usepackage{graphicx}

\usepackage{xcolor,pifont}
\newcommand*\colourcheck[1]{%
  \expandafter\newcommand\csname #1check\endcsname{\textcolor{#1}{\ding{51}}}%
}
\newcommand*\colouruncheck[1]{%
  \expandafter\newcommand\csname #1uncheck\endcsname{\textcolor{#1}{\ding{53}}}%
}
\colourcheck{green}
\colouruncheck{red}
\colouruncheck{blue}

\definecolor{mypink1}{rgb}{0.858, 0.188, 0.478}
\definecolor{mygreen}{rgb}{0.258, 0.888, 0.178}

\newcommand{\verify}[1]{\textcolor{black}{#1}}

\newcommand{\sz}[0]{\ddag}

\newcommand{\sk}[0]{\diamondsuit}
\newcommand{\sx}[0]{\spadesuit}
\newcommand{\sy}[0]{\clubsuit}

\newcommand{\sw}[0]{\heartsuit}

\title{Analyzing the Effectiveness of the Underlying Reasoning Tasks in Multi-hop Question Answering}

\author{
Xanh Ho,$^{\thefootnote{*} \; \sk, \sw}$
Anh-Khoa Duong Nguyen,$^{\thefootnote{*} \; \sy}$
Saku Sugawara,$^\sw$\and
Akiko Aizawa$^{\sk, \sw, \sx}$ \\
$^\sk$The Graduate University for Advanced Studies, Kanagawa, Japan\\
$^\sw$National Institute of Informatics, Tokyo, Japan \\
$^\sy$Independent Researcher \\
$^\sx$The University of Tokyo, Tokyo, Japan \\
{\tt \{xanh, saku, aizawa\}@nii.ac.jp} \\
{\tt dnanhkhoa@live.com} 
}

\begin{document}
\maketitle

\begingroup\def\thefootnote{*}\footnotetext{Equal contribution.}\endgroup

\begin{abstract}

To explain the predicted answers and evaluate the reasoning abilities of models, several studies have utilized underlying reasoning (UR) tasks in multi-hop question answering (QA) datasets. However, it remains an open question as to how effective UR tasks are for the QA task when training models on both tasks in an end-to-end manner. In this study, we address this question by analyzing the effectiveness of UR tasks (including both sentence-level and entity-level tasks) in three aspects: (1) QA performance, (2) reasoning shortcuts, and (3) robustness. While the previous models have not been explicitly trained on an entity-level reasoning prediction task, we build a multi-task model that performs three tasks together: sentence-level supporting facts prediction, entity-level reasoning prediction, and answer prediction. Experimental results on 2WikiMultiHopQA and HotpotQA-small datasets reveal that (1) UR tasks can improve QA performance. Using four debiased datasets that are newly created, we demonstrate that (2) UR tasks are helpful in preventing reasoning shortcuts in the multi-hop QA task. However, we find that (3) UR tasks do not contribute to improving the robustness of the model on adversarial questions, such as sub-questions and inverted questions. We encourage future studies to investigate the effectiveness of entity-level reasoning in the form of natural language questions (e.g., sub-question forms).\footnote{Our data and code are available at \url{https://github.com/Alab-NII/multi-hop-analysis}}


\end{abstract}

\section{Introduction}

\begin{figure*}[ht]
    \centering
    \includegraphics[width=\linewidth]{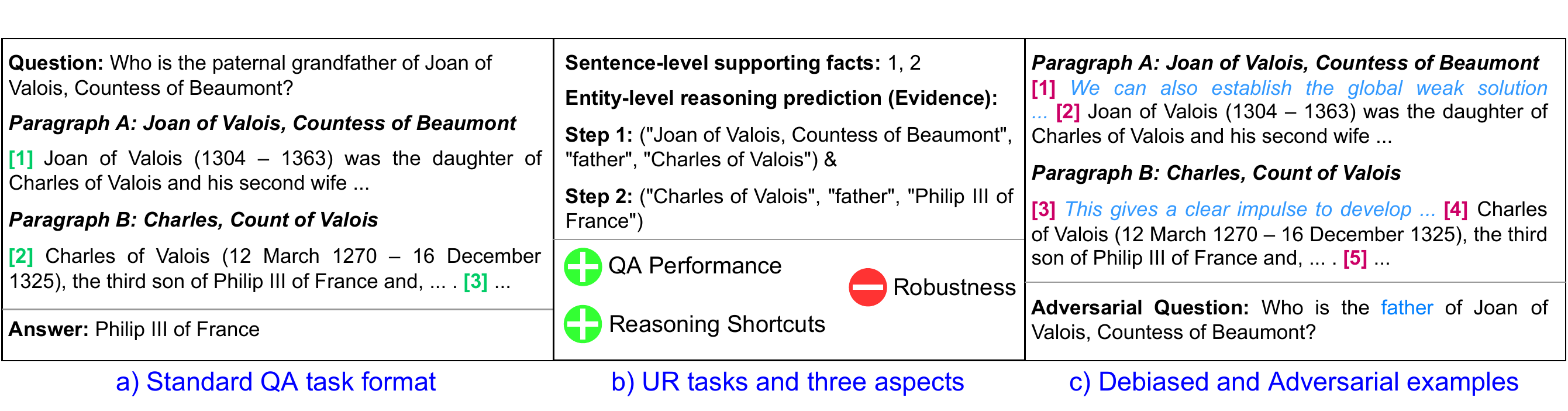}
    \caption{
    Example of (a) a standard multi-hop question, (b) two underlying reasoning tasks in the QA process and three aspects in our analysis, `+' and `-' indicate that the UR tasks have a positive and negative impacts, respectively, and (c) debiased and adversarial examples that are used in our study.
    } 
    \label{fig:example}
\end{figure*}

The task of multi-hop question answering (QA) requires a model to read and aggregate information from multiple paragraphs to answer a given question (Figure~\ref{fig:example}a).
Several multi-hop QA datasets have been proposed, such as QAngaroo~\cite{welbl-etal-2018-constructing}, HotpotQA~\cite{yang-etal-2018-hotpotqa}, and MuSiQue~\cite{TACL3639}.
In HotpotQA, the authors provide sentence-level supporting facts (SFs) to test the reasoning ability and explainability of the models.
However, owing to the design of the sentence-level SFs task (binary classification) and the redundant information in the sentences, \citet{inoue-etal-2020-r4c} and \citet{ho-etal-2020-constructing} show that the sentence-level SFs are insufficient to explain and evaluate multi-hop models in detail.
To address this issue, $\mathrm{R^4C}$~\cite{inoue-etal-2020-r4c} and 2WikiMultiHopQA~\cite[2Wiki;][]{ho-etal-2020-constructing} datasets provide an entity-level reasoning prediction task to explain and evaluate the process of answering questions.
Entity-level reasoning information is defined as a set of triples that
describes the reasoning path from question to answer (Figure~\ref{fig:example}b).

Several previous studies~\cite{chen2019multihop,fu-etal-2021-decomposing-complex} utilize sentence-level SFs and/or entity-level reasoning information to build explainable models by using question decomposition~\cite{min-etal-2019-multi,perez-etal-2020-unsupervised} or predicting sentence-level SFs. 
The advantages of these pipeline models are that they can exploit the underlying reasoning (UR) process in QA and their predicted answers are more interpretable.
However, the question remains as to how effective training on  UR tasks is for the QA task in an end-to-end manner.
Although a few end-to-end models have also been introduced~\cite{qiu-etal-2019-dynamically,fang-etal-2020-hierarchical}, these models are not explicitly trained on entity-level and answer prediction tasks.

In addition to the triple form, the sub-question form is another way to utilize entity-level reasoning information.
Specifically,~\citet{tang-etal-2021-multi} utilize question decomposition as an additional sub-question evaluation for bridge questions (there are two types of questions: bridge and comparison) in HotpotQA.
They only use sub-questions for evaluation and do not fine-tune the models on them. In addition, \citet{ho-etal-2022-well} use sub-questions for both evaluation and training. However, they only focus on comparison questions for date information.
In contrast, we focus on the triple form of the entity-level information and conduct experiments using two datasets, 2Wiki and HotpotQA-small (obtained by combining HotpotQA and $\mathrm{R^4C}$), which include both types of questions.

In this study, we analyze the effectiveness of UR tasks (including both sentence-level and entity-level) in three aspects: (1) \textit{QA performance}, (2) \textit{reasoning shortcuts}, and (3) \textit{robustness}.
First, QA performance is the final objective of the QA task. We aim to answer the following question:
\textbf{(RQ1)} \textit{Can the UR tasks improve QA performance?} 
For the second aspect, previous studies~\cite{chen-durrett-2019-understanding,jiang-bansal-2019-avoiding,min-etal-2019-compositional,trivedi-etal-2020-multihop} demonstrate that many questions in the multi-hop QA task contain biases and reasoning shortcuts~\cite{Geirhos_2020}, where the models can answer the questions by using heuristics.
Therefore, we aim to ask the following: 
\textbf{(RQ2)} \textit{Can the UR tasks prevent reasoning shortcuts?}
For the final aspect, to ensure safe development of NLP models, robustness is one of the important issues and has gained tremendous amount of research~\cite{wang-etal-2022-measure}. 
In this study, we aim to test the robustness of the model by asking modified versions of questions, such as sub-questions and inverted questions. Our question is
\textbf{(RQ3)} \textit{Do the UR tasks make the models more robust?}

There are no existing end-to-end models that can perform three tasks simultaneously (sentence-level SFs prediction, entity-level prediction, and answer prediction); therefore, we first build a multi-hop BigBird-base model~\cite{NEURIPS2020_c8512d14} to perform these three tasks simultaneously.
We then evaluate our model using two multi-hop datasets: 2Wiki and HotpotQA-small.
To investigate the effectiveness of the UR tasks, for each dataset, we conduct three additional experiments in which the model is trained on: (1) answer prediction task, (2) answer prediction and sentence-level prediction tasks, and (3) answer prediction and entity-level prediction tasks.
We also create four debiased sets (Figure~\ref{fig:example}c) for 2Wiki and HotpotQA-small for \textbf{RQ2}.
We create and reuse adversarial questions for 2Wiki and HotpotQA-small for \textbf{RQ3}.

The experimental results indicate that the UR tasks can improve QA performance from 77.9 to 79.4 F1 for 2Wiki and from 66.4 to 69.4 F1 for HotpotQA-small \textbf{(RQ1)}.
The results of the models on the four debiased sets reveal that the UR tasks can be used to reduce reasoning shortcuts \textbf{(RQ2)}.
Specifically, when the model is trained on both answer prediction and UR tasks, the performance drop of the model on the debiased sets is lower than that when the model is trained only on answer prediction (e.g., 8.9\%  vs. 13.4\% EM).
The results also suggest that the UR tasks do not make the model more robust on adversarial questions, such as sub-questions and inverted questions \textbf{(RQ3)}.
Our analysis shows that correct reconstruction of the entity-level reasoning task contributes to finding the correct answer in only 37.5\% of cases.
This implies that using entity-level reasoning information in the form of triples does not answer adversarial questions, in this case, the sub-questions.
We encourage future work to discover the effectiveness of the entity-level reasoning task in the form of sub-questions that have the same form as multi-hop QA questions.

\section{Background}
\label{sec_terminology}

\paragraph{Reasoning Tasks in Multi-hop QA}

In this study, we consider UR tasks
in multi-hop QA including two levels: \textit{sentence-level} and \textit{entity-level}.
The sentence-level SFs prediction task was first introduced by~\citet{yang-etal-2018-hotpotqa}.
This task requires a model to predict a set of sentences that is necessary to answer a question (Figure~\ref{fig:example}).

To evaluate the UR process of the models, derivation and evidence information were introduced in $\mathrm{R^4C}$ and 2Wiki, respectively.
Both derivation and evidence are sets of triples that represent the reasoning path from question to answer.
The difference is the form; derivation in $\mathrm{R^4C}$ uses a semi-structured natural language form, whereas evidence in 2Wiki uses a structured form.
We conduct experiments with both $\mathrm{R^4C}$ (HotpotQA-small) and 2Wiki. For consistency, we use the term \textit{entity-level reasoning prediction task} to denote the derivation task in $\mathrm{R^4C}$ and the evidence task in 2Wiki.

\paragraph{Reasoning Shortcuts and Biases}
\label{sec_position_bias}

In this study, we consider both reasoning shortcuts and biases to be similar.
These are spurious correlations in the dataset that allow a model to answer the question correctly without performing the expected reasoning skills, such as comparison and multi-hop reasoning.
Following previous studies~\cite{jiang-bansal-2019-avoiding,ko-etal-2020-look}, we use the terms \textit{word overlap shortcut} and \textit{position bias}.

To check whether the UR tasks can prevent reasoning shortcuts, we first identify the types of shortcuts that exist in HotpotQA-small and 2Wiki.
We use heuristics to identify the word overlap shortcut
(Appendix~\ref{app_word_overlap_shortcut}).
We find that the word overlap shortcut is common in HotpotQA-small, but not in 2Wiki.
The small sample size of HotpotQA-small (Section~\ref{sec_dataset}) increases the uncertainty of the obtained results.
Therefore, within the scope of this study, we mainly experiment with position bias.

\begin{figure}[t]
\centering
    \includegraphics[scale=0.450]{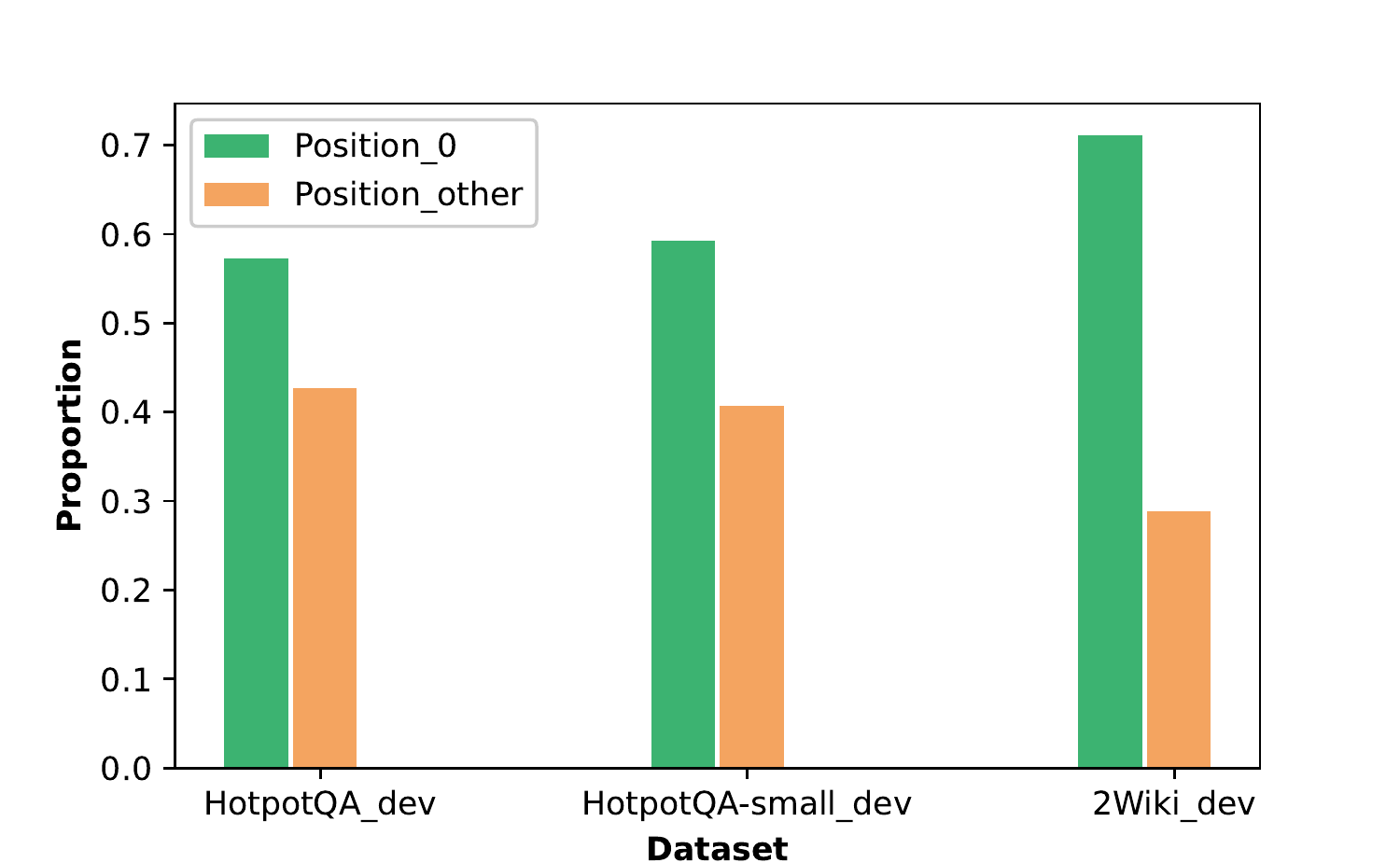}
    \caption{Information on the position of sentence-level SFs in the dev. sets of the three datasets.}
    \label{fig:bias_data}
\end{figure}

We observe that many examples in 2Wiki contain answers in the first sentence.
Therefore, we divide every sentence-level SF in each gold paragraph into two levels: the first sentence (position\_0) and the remaining sentences (position\_other).
Subsequently, we obtain the percentage of each level by dividing 
the total number of each level (e.g., position\_0) by the total number of SFs. 
Figure~\ref{fig:bias_data} illustrates the information on the position of sentence-level SFs in dev. sets of three datasets.
We find that all three datasets have a bias toward the first sentence.
We also find that 2Wiki has more position biases than HotpotQA and HotpotQA-small.


\section{Our Multi-task Model}
\label{sec_model}

\begin{figure}[t]
\centering
    \includegraphics[scale=0.65]{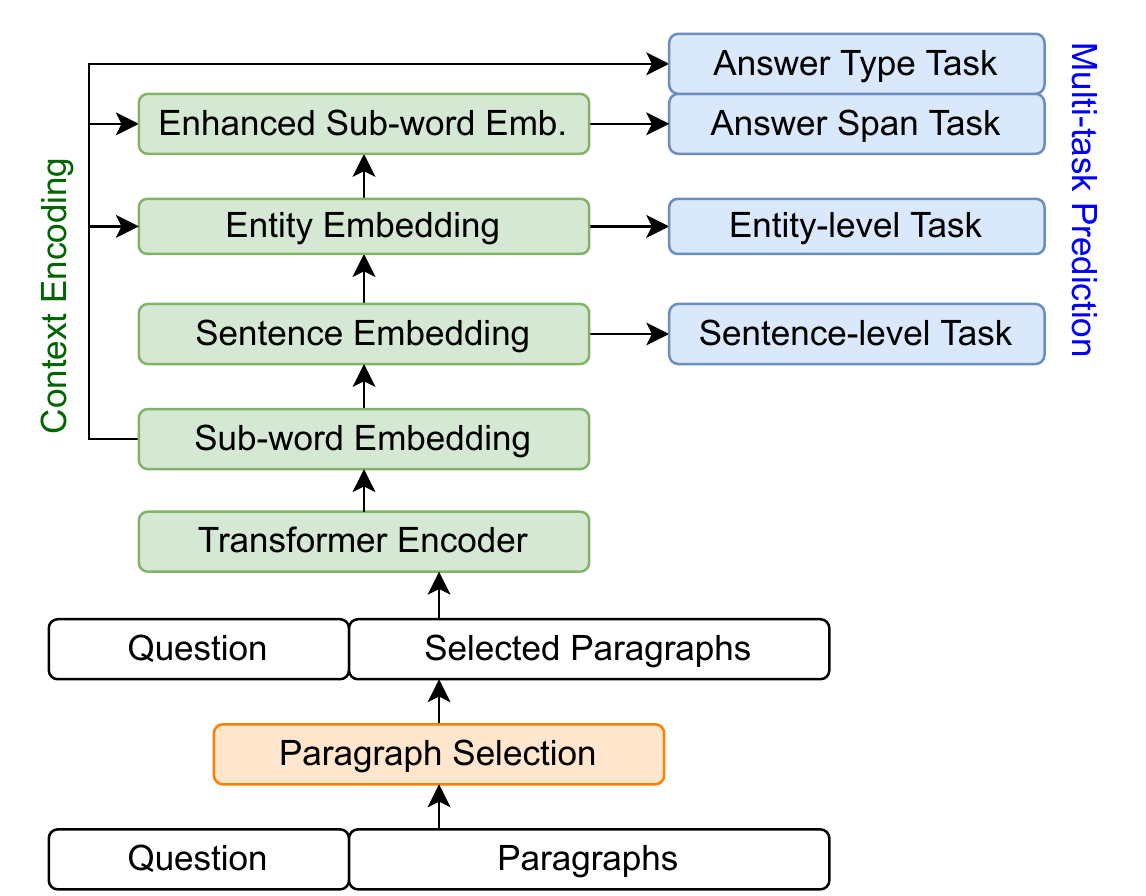}
    \caption{Our model has three main steps: paragraph selection, context encoding, and multi-task prediction.
    }
    \label{fig:model}
\end{figure}

To investigate the usefulness of UR tasks for the QA task, we jointly train the corresponding tasks: sentence-level SFs prediction, entity-level prediction, and answer prediction.
Figure~\ref{fig:model} illustrates our model.
To handle long texts, we use the BigBird model~\cite{NEURIPS2020_c8512d14}, which is available in Hugging Face's transformers repository.\footnote{\url{https://huggingface.co/transformers/model_doc/bigbird.html}}
Our model comprises three main steps: (1) paragraph selection, (2) context encoding, and (3) multi-task prediction.
We use the named entity recognition (NER) models of Spacy\footnote{\url{https://spacy.io/}} and Flair~\cite{akbik-etal-2019-flair} to extract all entities in the context and use them for the entity-level prediction task.

\renewcommand{\floatpagefraction}{.9}

\paragraph{Paragraph Selection}
Following previous models~\cite{qiu-etal-2019-dynamically,fang-etal-2020-hierarchical,tu2020sae}, instead of using all the provided paragraphs, we first filter out answer-unrelated paragraphs.
We follow the paragraph selection process described in~\citet{fang-etal-2020-hierarchical}.
First, we retrieve first-hop paragraphs by using title matching or entity matching.
We then retrieve second-hop paragraphs using the hyperlink information available in Wikipedia.
When we retrieve paragraphs, we reuse a paragraph ranker model\footnote{\url{https://github.com/yuwfan/HGN}} from the hierarchical graph network (HGN) model~\cite{fang-etal-2020-hierarchical} 
to rank input paragraphs using the probability of whether they contain sentence-level SFs.

\paragraph{Context Encoding}
To obtain vector representations for sentences and entities, we first combine all the selected paragraphs into one long paragraph and then concatenate it with the question to form a context $C$.
Specifically, $C = [\mathrm{[CLS]}, q_1, ..., q_m, \mathrm{[SEP]}, p_1, ..., p_n, \mathrm{[SEP]}]$, where 
$m$ and $n$ are the lengths of the question $q$ and the combined paragraph $p$ (all selected paragraphs), respectively.
The context $C$ is then tokenized into $l$ sub-words before feeding into BigBird to obtain the contextual representation $C'$ of the sub-words:
\begin{equation}
C' = \mathrm{BigBird}(C) \in \mathbb{R}^{l \times h} \mathrm{,}
\end{equation}
where $h$ is the hidden size of the BigBird model.
Next, we obtain the representation $s_i \in \mathbb{R}^{2h}$ of the $i$-th sentence and the representation $e_j \in \mathbb{R}^{4h + d_t}$ of the $j$-th entity, as follows:
\begin{equation}
\begin{aligned}
s_i = C'_{S_{\mathrm{start}}^i};C'_{S_{\mathrm{end}}^i} \\
e_j = C'_{E_{\mathrm{start}}^j};C'_{E_{\mathrm{end}}^j};t_j;s_k \mathrm{,}
\end{aligned}
\end{equation}
\noindent where [;] denotes the concatenation of the two vectors, 
$C'_{S_{\mathrm{start}}^i}$ and $C'_{E_{\mathrm{start}}^j}$ denote the first sub-word representations of the $i$-th sentence and $j$-th entity, respectively.
$C'_{S_{\mathrm{end}}^i}$ and $C'_{E_{\mathrm{end}}^j}$ denote the last sub-word representations of the $i$-th sentence and $j$-th entity, respectively.
We enrich the entity embedding $e_j$ by concatenating it with a $d_t$-dimensional type embedding $t_j$ and a sentence embedding $s_k$, where $k$ is the index of the sentence containing the $j$-th entity.

We also leverage the entity information to improve the contextual representation of sub-words $C'$ as it is mainly used for the answer prediction task, which will be described in the next section.
Thus, the enhanced sub-word representation $C''_i$ of the $i$-th sub-word is calculated as follows:
\begin{equation}
C''_i = C'_i;e_k \in \mathbb{R}^{5h + d_t} \mathrm{,}
\end{equation}
\noindent where $e_k$ is the embedding of the $k$-th entity containing the $i$-th sub-word. Otherwise, $e_k$ is a null vector with the same dimension.

\paragraph{Multi-task Prediction}
After context encoding, we train our model on three main tasks together: (1) sentence-level prediction, (2) entity-level prediction, (3) and answer prediction.
We split the answer prediction task into two sub-tasks, similar to previous studies~\cite{yang-etal-2018-hotpotqa,fang-etal-2020-hierarchical}, including answer type prediction and answer span prediction.
We train our model by minimizing the joint loss for all tasks, as follows:

\begin{equation}
\begin{aligned}
L_{\mathrm{joint}} = \lambda_{\mathrm{sent}}L_{\mathrm{sent}} + \lambda_{\mathrm{ent}}L_{\mathrm{ent}} +  \\ \lambda_{\mathrm{ans}}(L_{\mathrm{start}} + L_{\mathrm{end}} + L_{\mathrm{type}}) \mathrm{,}
\end{aligned}
\end{equation}
\noindent where $\lambda_{\mathrm{sent}}$, $\lambda_{\mathrm{ent}}$, and $\lambda_{\mathrm{ans}}$ are the hyper-parameters 
for three tasks: sentence-level prediction, entity-level prediction, and answer prediction (details are given in Appendix~\ref{app_detail_parameter}).

For the sentence-level prediction task, we use a binary classifier to predict whether a sentence is a supporting fact.
For the answer type prediction task, we use a 4-way classifier to predict the probabilities of \textit{yes}, \textit{no}, \textit{span}, and \textit{no answer}.
Two linear classifiers are used for the answer span prediction task to independently predict the start and end tokens of the answer span.

Different from existing end-to-end models~\cite{qiu-etal-2019-dynamically,fang-etal-2020-hierarchical}, our model is explicitly trained on the entity-level prediction task.
We formalize the entity-level reasoning prediction task as a relation extraction task~\cite{zhang2015relation}.
The input is a pair of entities, and the output is the relationship between two entities.
From all named entities obtained by using the NER models, we generate a set of entity pairs; for example, given $N$ entities, we obtain $N \times (N-1)$ pairs.
For each pair, we predict a relationship in a set of predefined relationships obtained from the training set.
 We then use cross-entropy as the learning objective.
%

\section{Datasets and Evaluation Metrics}
\label{sec_dataset}

We mainly experiment with 2Wiki and HotpotQA-small.
We also train and evaluate our model on the full version of HotpotQA.
We reuse and create debiased and adversarial sets for the evaluation.
Table~\ref{tab_dataset} presents the statistics for 2Wiki, HotpotQA-small, and additional evaluation sets.
The details of HotpotQA and 2Wiki are presented in Appendix~\ref{app_dataset}.
It should be noted that all datasets are in English.

\begin{table}[t]
  \begin{center}
    \begin{tabular}{l r r} 
    \toprule
 
    \textbf{Split}  & \textbf{2Wiki} & \textbf{HotpotQA-small}  \\

      \midrule
    Train   & 167,454 &   3,671\\
    
    Dev.  & 12,576 &  917 \\

    Test  & 12,576 & - \\
    
    Debiased  & 12,576 (x4)  &  917 (x4) \\  
    
    Adversarial  & 12,576  &  659 \& 134 \\   
      
    \bottomrule

    \end{tabular}
    \caption{
    Statistics for 2Wiki and HotpotQA-small. 
    There are four debiased sets in 2Wiki and HotpotQA-small. There are one adversarial set in 2Wiki and two adversarial sets in HotpotQA-small.
    }
    \label{tab_dataset}
  \end{center}
\end{table}

\subsection{HotpotQA-small}
\label{dataset_hotpot_small}

$\mathrm{R^4C}$~\cite{inoue-etal-2020-r4c} is created by adding entity-level reasoning information to the samples in HotpotQA.
We obtain HotpotQA-small by combining HotpotQA~\cite{yang-etal-2018-hotpotqa} with $\mathrm{R^4C}$.
HotpotQA-small comprises three tasks as in 2Wiki: (1) sentence-level SFs prediction, (2) entity-level prediction, and (3) answer prediction.
First, we re-split the ratio between the training and dev. sets; the new sizes are 3,671 and 917 for the training and dev. sets, respectively (the original sizes are 2,379 and 2,209, respectively).
In $\mathrm{R^4C}$, there are three gold annotations for the entity-level prediction task; in 2Wiki, there is only one gold annotation.
For consistency in the evaluation and analysis, we randomly choose one annotation from the three annotations for every sample in $\mathrm{R^4C}$.

The entity-level reasoning in $\mathrm{R^4C}$ is created by crowdsourcing.
We observe that there are many similar relations in the triples in $\mathrm{R^4C}$, and these relations can be grouped into one.
For example, \textit{is in}, \textit{is located in}, \textit{is in the}, and \textit{is located in the} indicate location relation.
We also group the relations by removing the context information in the relations; for example, \textit{is a 2015 book by} and \textit{is the second book by} are considered similar to the relation \textit{is a book by}.
After grouping, the number of relations in $\mathrm{R^4C}$ is 2,526 (it is 4,791 before).

\subsection{Debiased Dataset}

The objective of our debiased dataset is to introduce a small perturbation in each paragraph to mitigate a specific type of bias, in our case, the position bias shown in Figure~\ref{fig:bias_data}. 
For both 2Wiki and HotpotQA-small, we use the same method to generate four debiased sets: \textsc{AddUnrelated}, \textsc{AddRelated}, \textsc{Add2}, and \textsc{Add2Swap}.
The differences between these four sets are whether the sentence is related or unrelated to the paragraph and whether we add one or two sentences into the paragraph.
The details of each set are as follows.

\textit{\textbf{\textsc{AddUnrelated}}}: 
One sentence unrelated to the paragraph is added. 
    In our experiment, we use a list of sentences in the sentence-level revision dataset~\cite{tan-lee-2014-corpus}. 
    We randomly choose one sentence that has a number of tokens greater than eleven but less than twenty-one.

 \textit{\textbf{\textsc{AddRelated}}}: 
One sentence that does not have an impact on the meaning or flow of the paragraph is added. In our experiment, we write multiple templates for each entity type (e.g., for a film entity, ``\#Name is a nice film'', where \#\textit{Name} is the title of the paragraph), then randomly choose one template, and add it to the paragraph. %
    To detect the type of the paragraph, we use the question type information in 2Wiki and HotpotQA-small, the results of the NER model, and the important keywords in the question (e.g., who, magazine, album, and film).
    
\textit{\textbf{\textsc{Add2}}}:  \textsc{AddRelated} and \textsc{AddUnrelated} are combined in order.

\textit{\textbf{\textsc{Add2Swap}}}: The order of \textsc{AddRelated} and \textsc{AddUnrelated} in \textsc{Add2} is swapped.

\subsection{Adversarial Dataset}

The objective of our adversarial dataset is to check the robustness of the model by asking modified versions of questions.
For HotpotQA-small, we reuse two versions of adversarial examples in~\citet{geva2021break}.
The first one is automatically generated by using the `Break, Perturb, Build' (BPB) framework in~\citet{geva2021break}.
The BPB framework performs three main steps: (1) breaking a question into multiple reasoning steps, (2) perturbing the reasoning steps by using a list of defined rules, and (3) building new QA samples from the perturbations in step \#2.
The second version is a subset of the first version and is validated by crowd workers.
We only use the examples in these two versions that the original examples appear in HotpotQA-small.

For 2Wiki, no adversarial dataset is available.
Based on the idea of the BPB framework in~\citet{geva2021break}, we apply two main rules from BPB for 2Wiki: (1) replace the comparison operation for comparison questions, and (2) use the prune step for bridge questions.
For the first rule, we replace the operation in the comparison questions (e.g., ``Who was born first, A or B?'' is converted to ``Who was born later, A or B?'').
For the second rule, we use a sub-question in the QA process as the main question (e.g., for Figure~\ref{fig:example}, we ask,  ``Who is the father of Joan of Valois?'').

\begin{table*}[t]
  \begin{center}
    \begin{tabular}{l l r r r r r r} 
     \toprule
     
    \multirow{2}{2cm}{\textbf{Dataset}} &  \multirow{2}{2cm}{\textbf{Task Setting}} & \multicolumn{2}{c}{\textbf{Answer}}  & \multicolumn{2}{c}{\textbf{Sentence-level}}  &
      \multicolumn{2}{c}{\textbf{Entity-level}}  \\ 
      \cmidrule{3-8}
     &  & EM & F1 & EM & F1 & EM & F1 \\
       \midrule

     \multirow{4}{2cm}{2Wiki} & (1) Ans & 72.03 &	77.87  & -  &  - & - & - \\
       & (2) Ans + Sent &  72.82 &	78.65   &  78.06 &	92.38  & - & - \\
       & (3) Ans + Ent &  72.33	& 78.21   & -  & -  &  46.11 &	76.65 \\

       & (4) Ans + Sent + Ent &  \textbf{73.60} &	\textbf{79.37}   &   78.46 &	92.68   &  45.97	& 76.69 \\
       
         \midrule
       
     \multirow{4}{2cm}{HotpotQA-small} & (1) Ans & 52.89 &	66.43		  & -  &  - & - & - \\
       & (2) Ans + Sent &   54.42 &	69.03 &	75.35 &	91.00		  & - & - \\
       & (3) Ans + Ent &  54.74	 & 69.08 &	- &	- &	6.54 &	31.31	  	 \\

       & (4) Ans + Sent + Ent & \textbf{54.74} &	\textbf{69.44}	 & 75.14 &	90.88 &	6.43 &	31.05  	  \\

    \bottomrule

    \end{tabular}
    \caption{
    Ablation study results (\%) of our model in the dev. sets of 2Wiki and HotpotQA-small.
    \textit{Ans}, \textit{Sent}, and \textit{Ent} represent the answer prediction task, sentence-level SFs prediction task, and entity-level prediction task, respectively. 
    `Task Setting' represents the tasks that the model is trained on.
    `-' indicates the tasks the model is not trained on.
    }
    \label{tab_result_4setting_ver1}
  \end{center}
\end{table*}

\subsection{Evaluation Metrics}
Each task in HotpotQA and 2Wiki is evaluated by using two metrics: exact match (EM) and F1 score.
Following the evaluation script in HotpotQA and 2Wiki, we use joint EM and joint F1 to evaluate the entire capacity of the model.  
For HotpotQA, they are the products of the scores of two tasks: sentence-level prediction and answer prediction.
For 2Wiki and HotpotQA-small, they are the products of the scores of three tasks: sentence-level prediction, entity-level prediction, and answer prediction.

\section{Results}

Currently, there are no existing end-to-end models that explicitly train all three tasks together; therefore, in this study, we use our proposed model for analysis. 
We also compare our model with other previous models on the HotpotQA and 2Wiki datasets.
In general, the experimental results indicate that our model is comparable to previous models and can be used for further analyses.
We focus more on the analysis; therefore, the detailed results of the comparison are presented in Appendix~\ref{appendix_result_compare}.

\subsection{Effectiveness of the UR Tasks}
To investigate the effectiveness of the UR tasks, we train the model in four settings: (1) answer prediction only, (2) answer prediction and sentence-level SFs prediction, (3) answer prediction and entity-level prediction, and (4) all three tasks together.
%

\paragraph{QA Performance (RQ1)}

Our first research question is whether the UR tasks can improve QA performance.
To answer this question, we compare the results of different task settings described above.
The results are presented in Table~\ref{tab_result_4setting_ver1}.
For 2Wiki, using sentence-level and entity-level separately (settings \#2 and \#3), the QA performance does not change significantly.
The improvement is significant when we combine both the sentence-level and entity-level (setting \#4). 
Specifically, the scores when the model is trained on the answer prediction task only (setting \#1) and on both the answer prediction task and UR tasks (setting \#4) are 77.9 and 79.4 F1, respectively.
In contrast to 2Wiki, using sentence-level and entity-level separately, 
there is a larger QA performance improvement in HotpotQA-small.
Specifically, the F1 scores of settings \#2 and \#3 are 69.0 and 69.1, respectively, whereas, the F1 score of the first setting is 66.4.
Similar to 2Wiki, there is a large gap between the two settings, \#1 and \#4 (66.4 F1 and 69.4 F1, respectively).

In summary, these results indicate that both sentence-level and entity-level prediction tasks contribute to improving QA performance.
These results align with the findings in~\citet{yang-etal-2018-hotpotqa}, which shows that incorporating the sentence-level SFs prediction task can improve QA performance.
We also find that when combining both sentence-level and entity-level prediction tasks, the scores of the answer prediction task are the highest.

\begin{table*}[t]
  \begin{center}
  \resizebox{\textwidth}{!}{%
    \begin{tabular}{l l | r r | r r | r r | r r} 
    \toprule
   \multirow{3}{2.1cm}{\textbf{Dataset}} & \multirow{3}{2cm}{\textbf{Task Setting}} & \multicolumn{8}{c}{\textbf{Reduction (\%) on Four Debiased Sets}}  \\
    \cmidrule{3-10}

  &  &  \multicolumn{2}{c}{\textbf{\textsc{AddUnrelated}}} &  \multicolumn{2}{c}{\textbf{\textsc{AddRelated}}} &  \multicolumn{2}{c}{\textbf{\textsc{Add2}}} &  \multicolumn{2}{c}{\textbf{\textsc{Add2Swap}}} \\ 
  
  &  &  EM  & F1 &  EM  & F1 &  EM  & F1 &  EM  & F1 \\ 
  
    \midrule

\multirow{4}{1.5cm}{2Wiki} & (1) Ans & \underline{13.40} &	\underline{12.13}  & 	 3.55 &	3.46  & 	\underline{12.32}	& \underline{11.72}  & 	\underline{18.99} &	\underline{17.51}  \\ 

 & (2) Ans + Sent & 11.00	& 9.71  & 	\underline{4.16} &	\underline{4.22}  & 11.22 &	10.69   & 	17.62 &	16.24   \\ 

 & (3) Ans + Ent & \textbf{7.73} &	\textbf{6.94} & 	\textbf{2.80} &	\textbf{2.77}   & \textbf{8.38} &	\textbf{7.76}  & \textbf{13.12} &	\textbf{12.21}  \\ 

 & (4) Ans + Sent + Ent &  8.86 &	8.11  & 	 3.16	 & 3.13   & 	9.09	 & 8.58   & 14.53 &	13.77    \\ 

\midrule

\multirow{4}{1.5cm}{HotpotQA-small} & (1) Ans &  3.01 &	1.53	  & 	  \underline{4.04}	 & \underline{1.50}	  & 		1.65 &	1.01  &  3.96 &	2.47	  \\ 

 & (2) Ans + Sent &  \textbf{1.13} &	\textbf{1.35}	  & 	  -0.51 &	0.19  & 	\textbf{0.08} &	\textbf{0.85}   &   \textbf{1.77} &	\textbf{1.96}	  \\ 

& (3) Ans + Ent & \underline{6.73} &	\underline{5.60}	  & 	 \textbf{-0.92} &	\textbf{0.03}	  & 	\underline{4.02} &	\underline{3.54}   &  \underline{6.89} &	\underline{5.46}  \\ 

 & (4) Ans + Sent + Ent &  5.05	& 4.65	  & 	1.26	 & 1.25	  & 	 1.83 &	2.46   &   3.58 &	3.64	  \\

     \bottomrule
    \end{tabular}}
    \caption{
  Average performance drop from five times running (smaller is better) of the four settings on the four debiased sets of 2Wiki and HotpotQA-small. The best and worst scores are boldfaced and underlined, respectively.
    }
    \label{tab_result_adversarial_both}
  \end{center}
\end{table*}

\paragraph{Reasoning Shortcuts (RQ2)}

To investigate whether explicitly optimizing the model on the UR tasks can prevent reasoning shortcuts, we evaluate the four settings of the model on the four debiased sets of 2Wiki and HotpotQA-small.
The generation of the debiased sets includes stochastic steps. To minimize the impact of randomness on our reported results, we generate the debiased sets five times and report the average evaluation scores.
The average performance drops are presented in Table~\ref{tab_result_adversarial_both}
(detailed scores are given in Appendix~\ref{r_ques_2}).

%
Overall, for 2Wiki, when the model is trained on only one task (\#1), the drop is the largest (except for \textsc{\textsc{AddRelated}}, which is the second largest).
When the model is trained only on the answer prediction task, the drops are always higher than those when the model is trained on three tasks.
Specifically, the gaps between the two settings, \#1 (only answer task) and \#4 (all three tasks), are 4.5\%, 0.4\%, 3.2\%, 4.5\% (EM score) for \textsc{AddUnrelated}, \textsc{AddRelated}, \textsc{Add2}, and \textsc{Add2Swap}, respectively.
These scores indicate that the two tasks, sentence-level and entity-level, positively affect the answer prediction task when the model is trained on three tasks simultaneously.

For HotpotQA-small, we observe that the effectiveness of the UR tasks is inconsistent.
For example, for \textsc{AddUnrelated}, when training the model on the three tasks (setting \#4), the reduction is larger than that when training on answer task only (setting \#1) (5.1 vs. 3.0 EM).
However, for \textsc{AddRelated}, the reduction on setting \#4 is smaller than that on setting \#1 (1.3 vs. 4.0 EM).
One possible reason is that the performance of the entity-level task is not good (6.4 EM), which affects the answer prediction task when the model is trained on the three tasks together.
Another possible reason is that the position bias in HotpotQA-small is not sufficiently large.
We present a detailed analysis in Section~\ref{analyses_sec} to explain this case.

\paragraph{Robustness (RQ3)}

\begin{table}[t]
  \begin{center}
 \resizebox{\columnwidth}{!}{%
    \begin{tabular}{l  r r r r } 
     \toprule
     
 \multirow{2}{1.99cm}{\textbf{Task Setting}} &  \multicolumn{2}{c}{\textbf{Dev-adver}} & \multicolumn{2}{c}{\textbf{Reduction \%}}  \\ 
      \cmidrule{2-5}
    &  EM & F1  & EM & F1 \\
       \midrule

     Ans  & 37.09  & 46.07    & \textbf{48.51} &	\textbf{40.84} \\
     
       Ans + Sent &   34.26 & 43.64  & 52.95 &	44.51  \\
       
      Ans + Ent   & 32.67  & 39.43  & 54.83	& 49.58 \\
       
    Ans + Sent + Ent &  34.19 & 42.74  &  53.55	& 46.15	 \\
    
    \bottomrule

    \end{tabular}
      }
    \caption{Results of our model in the dev-adversarial set of 2Wiki and the performance drop.
    }
    \label{tab:result_4setting_adver}
  \end{center}

\end{table}

To test whether the UR tasks can help to improve the robustness of the model, we evaluate the four settings of the model on the adversarial sets.
For 2Wiki, the results are presented in Table~\ref{tab:result_4setting_adver}.
The scores for all four settings decrease significantly on the adversarial set.
The reduction is the smallest when the model is trained on the answer task only.
The UR tasks do not make the model more robust on this adversarial set.
For HotpotQA-small, we observe the same behavior, that is, when the model is trained on the answer task only, the reduction is the smallest.
All results are presented in Table~\ref{tab:result_2setting_adver_hotpot_ver2}.
These results indicate that both sentence-level and entity-level prediction tasks do not contribute to improving the robustness of the models on adversarial questions, such as sub-questions and inverted questions.
We analyze the results in Section~\ref{detail_ana_rq3}.

\begin{table*}[t]
  \begin{center}
 \resizebox{\textwidth}{!}{%
    \begin{tabular}{l r r | r r  r r | r r r r} 
     \toprule
     
 \multirow{2}{1.99cm}{\textbf{Task Setting}} & \multicolumn{2}{c}{\textbf{Dev}}  & \multicolumn{2}{c}{\textbf{Dev-Adver}}  & \multicolumn{2}{c}{\textbf{Adver$\downarrow$} (\%)}  
     & \multicolumn{2}{c}{\textbf{Dev-Adver-val}}      & \multicolumn{2}{c}{\textbf{Adver-val$\downarrow$} (\%)} \\ 
      \cmidrule{2-11}
     & EM & F1 & EM & F1 & EM & F1 & EM & F1 & EM & F1 \\
       \midrule

(1)  Ans &  52.89 &	66.43	  &   40.36 &	51.23 &	23.69 &	\textbf{22.88}   &  37.31 &	46.69 &	\textbf{29.46} &	\textbf{29.72}	 \\
 
(2) Ans + Sent &  54.42 &	69.03	  &  41.73 &	52.50 &	23.32 &	23.95   &  34.33 &	43.86 &	36.92 &	36.46		 \\
 
(3) Ans + Ent  &  54.74 &	69.08	  &  42.79 &	52.16 &	\textbf{21.83} &	24.49  &   27.61 &	36.86 &	49.56 &	46.64		 \\
 
(4) Ans + Sent + Ent &  54.74 &	69.44  &   40.52 &	51.14 &	25.98 &	26.35   &   31.34 &	38.22 &	42.75 &	44.96		 \\

    \bottomrule

    \end{tabular}
      }
    \caption{Results of our model in the dev. and two dev-adversarial sets of HotpotQA-small. `Adver' denotes adversarial and `Adver-val' denotes the adversarial set that was validated by crowd workers.
    %
    }
    \label{tab:result_2setting_adver_hotpot_ver2}
  \end{center}

\end{table*}

\begin{table}[t]
  \begin{center}
 \resizebox{\columnwidth}{!}{%
    \begin{tabular}{l  c  c  c} 
     \toprule
     
 \multirow{2}{1.99cm}{\textbf{Task Setting}} &  
  \multirow{2}{1.09cm}{\textbf{Correct Ans}} & 
  \multirow{2}{1.09cm}{\textbf{Correct Ent}} &
   \multirow{2}{2.19cm}{\textbf{Correct Both Ans \& Ent }} \\
   
      
    &  &  &   \\
       \midrule

     (3) Ans + Ent   & 4,109	& 6,851 & 2,249 (32.8\%) \\
       
   (4) Ans + Sent + Ent &   4,300	& 6,450 & 2,420 (37.5\%) \\
    
    \bottomrule

    \end{tabular}
      }
    \caption{
    Number of correct predicted answers, number of correct predicted entity-level reasoning, and number of examples that have both correct predicted answers and correct predicted entity-level reasoning.
    }
    \label{tab:analysis_rq3}
  \end{center}

\end{table}

\subsection{Analyses}
\label{analyses_sec}

\paragraph{Details of RQ2}

To investigate the results concerning RQ2 in more depth, we first analyze the position biases of different types of questions in 2Wiki and HotpotQA-small.
We find that the comparison questions have more position biases than the bridge questions in both 2Wiki and HotpotQA-small
(Appendix~\ref{app_position_bias}).
To evaluate the effectiveness of the position bias for each type of question, we evaluate the four settings of the model on the four debiased sets for each type of question in both datasets.
All the results are presented in Appendix~\ref{analysis_app}.

For 2Wiki, we find that most of the answers are in the first sentences in the comparison questions. This large bias is the main reason for the significant reduction in the scores in the comparison questions.
2Wiki has 46.0\% of comparison questions.
The reduction in comparison questions contributes to the reduction in the entire dataset.
In other words, the results of 2Wiki are affected by those of the comparison questions.
HotpotQA-small has only 22.0\% of comparison questions, and the position bias in the comparison questions was not sufficiently large.
Therefore, the position bias does not have a significant impact on the main QA task.
In other words, the UR tasks do not have a significant effect.

\paragraph{Details of RQ3}
\label{detail_ana_rq3}

The adversarial questions used in RQ3 are the sub-questions in the QA process for bridge questions and the inverted questions for comparison questions.
We observe that the triple in the entity-level task is helpful in answering the sub-questions.
For example, the triple is: (\textit{Charles of Valois, father, Philip III of France}) and the sub-question is ``\textit{Who is the father of Charles of Valois?}''.
To understand more on the behaviors of the model, we analyze the results from 2Wiki in two settings: (3) Ans + Ent and (4) Ans + Sent + Ent. 
Table~\ref{tab:analysis_rq3} presents the detailed results for these two settings.
We find that correct reconstruction of the entity-level reasoning task contributes to finding the correct answer only in 32.8\% of cases in setting \#3 and only in 37.5\% of cases in setting \#4.
Entity-level reasoning in the form of triples has no significant effect on the main QA process.
Several examples are presented in Appendix~\ref{analysis_app}.

We conjecture that there are three possible reasons why the UR tasks cannot contribute to the adversarial dataset.
The first one is the difference in the form and design of the tasks. 
Specifically, the entity-level reasoning task is formulated as a relation extraction task; the input is a pair of entities, and the output is a relation label. Meanwhile, the adversarial dataset is formulated as a QA task; the input is a natural language question, and the output is an answer.
The second reason is the incompetence of the entity-level reasoning information.
As discussed in~\citet{ho-etal-2022-well}, the entity-level reasoning in the comparison questions does not describe the full path from question to answer, and other reasoning operations are required to obtain the answer.
The final reason is the manner in which we utilize the entity-level reasoning information. 
Our model does not consider the order of the triples in the reasoning chain. For example, we do not consider the order of the two steps in Figure~\ref{fig:example}b.
We hope that our research will inspire future studies to investigate the effectiveness of the UR tasks in the form of a natural language question, which has the same form as a multi-hop QA question.

\section{Related Work}

\paragraph{Multi-hop Datasets and Analyses}

To test the reasoning abilities of the models, many multi-hop QA datasets~\cite{welbl-etal-2018-constructing, talmor-berant-2018-web, yang-etal-2018-hotpotqa} have been proposed.
Recently,~\citet{TACL3639} introduced MuSiQue, a multi-hop dataset constructed from a composition of single-hop questions. 
The reason why do we not conduct experiments on MuSiQue is explained in the limitations section.

In addition to \citet{tang-etal-2021-multi} and \citet{ho-etal-2022-well}, the most similar to our research mentioned in the Introduction,
there are some other existing studies~\cite{chen-durrett-2019-understanding,jiang-bansal-2019-avoiding,min-etal-2019-compositional,trivedi-etal-2020-multihop} on the analysis and investigation of the multi-hop datasets and models.
However, most of them do not utilize the internal reasoning information when answering questions.

\paragraph{Multi-hop Models}
Various directions have been proposed for solving multi-hop datasets, including question decomposition~\cite{talmor-berant-2018-web,jiang-bansal-2019-self,min-etal-2019-multi,perez-etal-2020-unsupervised,wolfson-etal-2020-break,fu-etal-2021-decomposing-complex}, iterative retrieval~\cite{feldman-el-yaniv-2019-multi,asai2020learning,qi-etal-2021-answering}, graph neural networks~\cite{song2018exploring,de-cao-etal-2019-question,ding-etal-2019-cognitive,qiu-etal-2019-dynamically,tu-etal-2019-multi,fang-etal-2020-hierarchical}, and other approaches such as single-hop based models~\cite{yang-etal-2018-hotpotqa,nishida-etal-2019-answering} or transformer-based models~\cite{devlin-etal-2019-bert,NEURIPS2020_c8512d14}. 
Our model is based on the BigBird transformer model.

\paragraph{Other QA Reasoning Datasets}
In addition to multi-hop reasoning datasets, several other existing datasets also aim to evaluate the reasoning abilities of the models. Some of them are: DROP~\cite{dua-etal-2019-drop} for numerical reasoning; CLUTRR~\cite{sinha-etal-2019-clutrr}, ReClor~\cite{Yu2020ReClor}, and LogiQA~\cite{10.5555/3491440.3491941} for logical reasoning;  Quoref~\cite{dasigi-etal-2019-quoref} for coreference reasoning; CommonsenseQA~\cite{talmor-etal-2019-commonsenseqa}, MCScript2.0~\cite{ostermann-etal-2019-mcscript2}, and CosmosQA~\cite{huang-etal-2019-cosmos} for commonsense reasoning. 
Many of these datasets consist of only a single paragraph in the input or lack explanation information that describes the reasoning process from question to answer.
However, our focus is on multi-hop reasoning datasets that contain multiple paragraphs in the input and provide explanatory information for the QA process.

\section{Conclusion}

We analyze the effectiveness of the underlying reasoning tasks using two multi-hop datasets: 2Wiki and HotpotQA-small.
The results reveal that the underlying reasoning tasks can improve QA performance.
Using four debiased sets, we demonstrate that the underlying reasoning tasks can reduce the reasoning shortcuts of the QA task.
The results also reveal that the underlying reasoning tasks do not make the models more robust on adversarial examples, such as sub-questions and inverted questions.
We encourage future studies to investigate the effectiveness of the entity-level reasoning task in the form of sub-questions.


\section*{Limitations}

Our study has two main limitations.
The first one is the small size of HotpotQA-small.
Currently, there are no other multi-hop datasets that contain a large number of examples with the entity-level reasoning prediction task.
MuSiQue is the most potential option.
The entity-level reasoning information in MuSiQue includes two types of formats: triple format and natural language question format.
We do not experiment with MuSiQue because the number of examples that have entity-level reasoning information in the form of a triple is small: 2,253 out of 19,938 in the training set and 212 out of 2,417 in the dev. set.

The second limitation is that our model does not consider the order of the triples in the entity-level reasoning prediction task.
As shown in Figure~\ref{fig:example}b, the two triples are ordered.
However, our model formulizes the entity-level prediction task as a relation extraction task. We predict a relation given the two entities detected by the NER models.
Therefore, the order of the triples is not considered.
We conjecture that this may be one of the reasons why the entity-level reasoning prediction task (e.g., a triple \textit{(Film A, director, D)}) does not support the model when answering sub-questions (e.g., \textit{Who is the director of Film A?}) using the same information.

\section*{Acknowledgments}
We would like to thank Viktor Schlegel and the anonymous reviewers for their valuable comments and suggestions. 
This work was supported by JSPS KAKENHI Grant Numbers 21H03502 and 22K17954 and JST AIP Trilateral AI Research and PRESTO Grant Number JPMJCR20G9.

\bibliography{anthology,custom}
\bibliographystyle{acl_natbib}

\appendix

\section{Word Overlap Shortcut}
\label{app_word_overlap_shortcut}

Using adversarial methods, \citet{jiang-bansal-2019-avoiding} show that examples in HotpotQA often contain word overlap shortcut, where the models can answer the questions by performing word-matching between the question and a sentence in the context.

Based on this finding, we automatically calculate the word overlap shortcut for 2Wiki and HotpotQA-small.
We observe that the word overlap shortcut is common in bridge questions; therefore, we only calculate the word overlap shortcut for bridge questions in 2Wiki and HotpotQA-small. 
To check whether a sample contains the word overlap shortcut, we do the following steps:

\begin{itemize}
    \item Obtain a set of surrounding words $S$ by getting the five words immediately to the left and right of the answer span, then remove stopwords in $S$.
    
    \item Obtain a set of overlapping words ($O$) between $S$ and a question.
    
    \item We consider a sample containing the word overlap shortcut if there are at least two words in $O$ and $\frac{|O|}{|S|} \geq 0.65$. These numbers (threshold) are chosen based on the evaluation of 40 examples that are manually annotated by the authors.
    
\end{itemize}

We find that there are 56 out of 5,791 and 151 out of 715 examples (5,791 and 715 are the numbers of bridge questions in 2Wiki and HotpotQA-small) in the dev. sets of 2Wiki and HotpotQA-small containing the word overlap shortcut.

It is noted that there is another type of shortcut, namely, entity-type matching shortcut.
Based on the experimental results and human performance,~\citet{min-etal-2019-compositional} reveal that examples in HotpotQA contain the entity type matching shortcut, where the models can answer the questions by using the first five tokens in the questions; meanwhile, humans can answer the questions by using the entity type of the paragraphs. 
Currently, there is no dataset that can prevent the entity-type shortcut; therefore, we do not use this type of shortcut in our experiments.

\section{Experimental Details}
\label{sec:appendix1}

\subsection{Implementation Details}
\label{app_detail_parameter}

We use Pytorch~\cite{paszke2017automatic} and Hugging Face when building our model.
For the context encoding step, we use a pre-trained BigBird model as the encoder; the hidden dimension is 768.
For the entity-level reasoning prediction task, we obtain 33 relations for 2Wiki and 2,526 relations for $\mathrm{R^4C}$, from all triples in the training set, including a non-relation type.
We use entity type embedding $d_t$ of 50.
We fine-tuned our model with a total batch size of 32 on a single GPU (NVIDIA A100 80GB) using mixed precision and a gradient accumulation step of 8.
Following the hyperparameters in the BERT model~\cite{devlin-etal-2019-bert}, for optimization, we use the Adam Opitmizer~\cite{DBLP:journals/corr/KingmaB14} with a learning rate of 3e-5, weight decay of 0.01, learning rate warmup over the first 10\% of the total number of training steps, and linear decay of the learning rate.
We also use a dropout probability of 0.1 on all layers.

For multi-task prediction, we use $\lambda_{\mathrm{sent}}$ as 4, $\lambda_{\mathrm{ent}}$ as 15, and $\lambda_{\mathrm{ans}}$ as 1 for 2Wiki and HotpotQA-small; we use $\lambda_{\mathrm{sent}}$ as 7 and $\lambda_{\mathrm{ans}}$ as 1 for HotpotQA.
We do not run all experiments with different values of $\lambda_{\mathrm{sent}}$, $\lambda_{\mathrm{ent}}$, and $\lambda_{\mathrm{ans}}$; instead we run several experiments, base on the results, we then adjust the parameters.
We find that when running with $\lambda_{\mathrm{sent}}$ as 4 for 2Wiki and 7 for HotpotQA, $\lambda_{\mathrm{ent}}$ as 15, and $\lambda_{\mathrm{ans}}$ as 1, it produces the best results.
We fix the random seed for the reproducibility of the results.
We observe that the final epoch often produces the best scores, and its scores are stable on adversarial datasets; therefore, 
we choose the final epoch for all settings in our experiment.

\subsection{Datasets}
\label{app_dataset}

\paragraph{HotpotQA}
HotpotQA was created by crowdsourcing.
Due to the design of the dataset, there are only two tasks in HotpotQA: sentence-level SFs prediction and answer prediction.
$\mathrm{R^4C}$ was created based on HotpotQA and contained 4,588 questions. 
The dataset requires systems to provide an answer and derivation in a semi-structured natural language form.
There are two types of questions in HotpotQA: bridge and comparison.

\begin{figure}[t]
    \centering
    \includegraphics[scale=0.695]{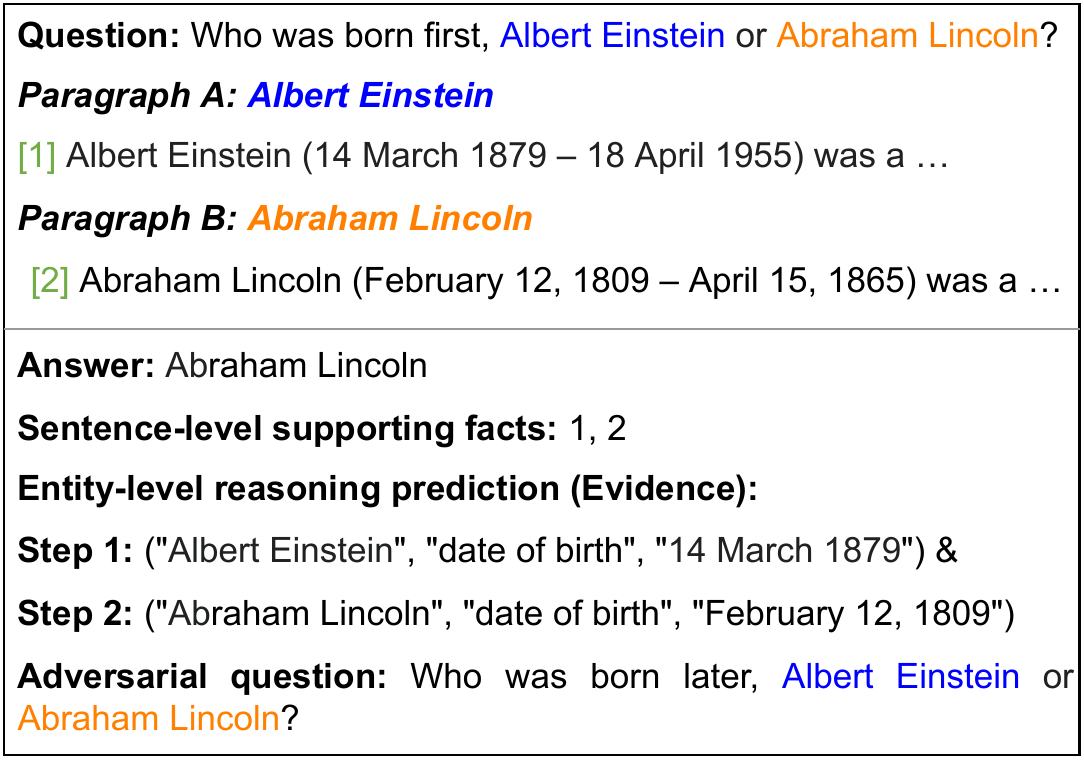}
    \caption{Example of a comparison question from the 2Wiki dataset.
    } 
    \label{fig_example_compare}
\end{figure}

\paragraph{2Wiki}
2Wiki was constructed by utilizing a Knowledge Base and Wikipedia, and the questions were created by using templates.
There are three different tasks in the dataset: (1) sentence-level SFs prediction, (2) evidence generation (for consistency, we use the term \textit{entity-level prediction}), and (3) answer prediction.
The context consists of ten paragraphs, including two or four gold paragraphs and eight or six distractor paragraphs.
The gold paragraph contains the information required to find the answer.
Meanwhile, the purpose of the distractor paragraph is to distract the models.
There are four different types of questions in the dataset: comparison, inference, compositional, and bridge-comparison.
Inference and compositional questions are two sub-types of the bridge question.
For the convenience of analysis, we consider comparison and bridge-comparison questions as comparison questions.
Figure~\ref{fig_example_compare} presents an example of a comparison question from the 2Wiki dataset.

2Wiki was designed to focus on the entire reasoning process from question to answer.
The entire capacity of the model is evaluated by using two metrics: joint EM and joint F1. 
To obtain the joint F1 score, they first calculate the joint precision and joint recall as follows: $P^{joint} = P^{ans}P^{ent}P^{sent}$ and $R^{joint} = R^{ans}R^{ent}R^{sent}$.
($P^{ans}, R^{ans}$), ($P^{ent}, R^{ent}$), ($P^{sent}, R^{sent}$) represent the precision and recall for three tasks: answer prediction, entity-level reasoning  prediction, and sentence-level SFs prediction.
The joint EM is 1 when all three tasks achieve an exact match and otherwise 0.

\subsection{Results Comparison}
\label{appendix_result_compare}

We compare our results with three previous models: BiDAF, CRERC, and NA-Reviewer.
BiDAF is a baseline model in~\citet{ho-etal-2020-constructing}.
CRERC~\cite{fu-etal-2021-decomposing-complex} is a pipeline model that includes three modules: relation extractor, reader, and comparator.
NA-Reviewer~\cite{article_fu} is an improved version of CRERC, as it addresses the error accumulation issue.
It is noted that both CRERC and NA-Reviewer models are evaluated on only 2Wiki.

%

Table~\ref{tab:result} presents the results of our model and previous models in the dev. set of HotpotQA and in the test set of 2Wiki.
It also shows the performance of our model in the dev. set of HotpotQA-small and human performance in~\citet{ho-etal-2020-constructing}.

\begin{table*}[t]
  \begin{center}
    \resizebox{\textwidth}{!}{%
    \begin{tabular}{l l r r r r r r r r} 
     \toprule
     
     \multirow{2}{1cm}{\textbf{Dataset}} &
      \multirow{2}{2cm}{\textbf{Model}} & \multicolumn{2}{c}{\textbf{Answer}}  & \multicolumn{2}{c}{\textbf{Sentence-level}}  &
      \multicolumn{2}{c}{\textbf{Entity-level}} & 
      \multicolumn{2}{c}{\textbf{Joint}} \\ 
      
      \cmidrule{3-10}
     &  & EM & F1 & EM & F1 & EM & F1 & EM & F1 \\
       \midrule
       
  \multirow{5}{1cm}{HotpotQA}  &  HGN-BERT$^\sz$~\cite{fang-etal-2020-hierarchical} & \textit{N/A} & 74.76   & \textit{N/A} & 86.61 & \blueuncheck & \blueuncheck & \textit{N/A} & 66.90  \\

    &  HGN-RoBERTa~\cite{fang-etal-2020-hierarchical} & \textbf{68.93} & \textbf{82.18}  & 63.09  & \textbf{88.59} & \blueuncheck & \blueuncheck & 46.46 & \textbf{74.34}  \\

    &  SAE-BERT~\cite{tu2020sae} & 61.32 & 74.81  & 58.06  & 85.27 & \blueuncheck & \blueuncheck & 39.89 & 66.45  \\ 
           
    &  SAE-RoBERTa~\cite{tu2020sae} & 67.70  & 80.75  & \textbf{63.30}  & 87.38 & \blueuncheck & \blueuncheck & \textbf{46.81} & 72.75  \\

     \cmidrule{2-10}

   & Our BigBird-base & 61.90 &  76.09 & 58.54  & 86.93 & \blueuncheck & \blueuncheck & 39.39 &  67.81 \\       

    \midrule

 HotpotQA-small &  Our BigBird-base  & 54.74 &  69.44 &  75.14 & 90.88 & 6.43 & 31.05 & 4.25 & 21.69  \\      

 \midrule
 
   \multirow{5}{1cm}{2Wiki}  &      BiDAF~\cite{ho-etal-2020-constructing}  & 36.53 & 43.93 & 24.99 & 65.26 & \hphantom{0}1.07 & 14.94 & \hphantom{0}0.35 &  \hphantom{0}5.41 \\ 
       
    &   CRERC~\cite{fu-etal-2021-decomposing-complex} & 69.58 &  72.33 &  82.86 &  90.68 & \textbf{54.86} & 68.83 &  49.80 & 58.99 \\

  &  NA-Reviewer~\cite{article_fu} & \textbf{76.73} &  \textbf{81.91} &  \textbf{89.61} &  \textbf{94.31} & 53.66 & 70.83 &  \textbf{52.75} & \textbf{65.23} \\ 
       \cmidrule{2-10}

  &  Our BigBird-base  & 74.05 & 79.68 & 77.14 & 92.13 & 45.75 & \textbf{76.64} & 39.30 & 63.24  \\

    \cmidrule{2-10}
    & Human UB~\cite{ho-etal-2020-constructing} & 91.00 & 91.79 & 88.00 & 93.75 & 64.00  & 78.81  & 62.00 & 75.25 \\      

    \bottomrule
    \end{tabular}%
    }
    \caption{
    Results (\%) of our model and previous models in the dev. set of HotpotQA and in the test set of 2Wiki.
    We also show the performance of our model in the dev. set of HotpotQA-small.
    \textit{Answer}, \textit{Sentence-level}, and \textit{Entity-level} represent the answer prediction task, sentence-level prediction task, and entity-level prediction task, respectively. 
    For HGN-BERT, the scores that we obtained (from left to right: 58.93 73.18 54.64 85.34 35.11 64.24) are lower than the reported scores in HGN~\cite{fang-etal-2020-hierarchical}; therefore, we show the reported F1 scores in HGN.
    }
    \label{tab:result}
  \end{center}
\end{table*}

\paragraph{Results on HotpotQA}
%
Our score is comparable to the BERT-base version of two strong models, SAE~\cite{tu2020sae} and HGN~\cite{fang-etal-2020-hierarchical} in the dev. set of the distractor setting in HotpotQA.
Specifically, our joint F1 is 67.8, while 
for SAE-BERT, it is 66.5, and for HGN-BERT, it is 66.9.
However, our score is smaller than the RoBERTa-base of SAE and HGN. 
They are 72.8 and 74.4 F1 for SAE-RoBERTa and HGN-RoBERTa, respectively.
It is noted that we use the BigBird-ITC version in our model.
Although the BigBird-ETC version performs better than the BigBird-ITC version, it is not available in Hugging Face.
We do not use SAE and HGN for our analyses because these models are not designed to train on the entity-level reasoning prediction task.

\paragraph{Results on HotpotQA-small}
The scores on HotpotQA-small are lower than those on HotpotQA in the answer prediction task.
This result may be explained by the fact that the training size of HotpotQA-small is smaller than HotpotQA (3,671 vs. 90,564). 
Due to the small size, we only use the gold paragraphs for experiments.
That is why the scores on HotpotQA-small are higher than those on HotpotQA in the sentence-level task.
For the entity-level task, the EM score is quite low (6.4 EM).
A possible reason for this is that there are many relations in HotpotQA-small (2,526 relations);
meanwhile, there are only 33 relations in 2Wiki.
We observe that the F1 score (31.1 F1) is much better than the EM score.
Therefore, we keep using HotpotQA-small for analyses.

\paragraph{Results on 2Wiki}

Our model significantly outperforms BiDAF in all tasks.
Our results are comparable to CRERC.
The EM score of our model in the entity-level task is lower than that of CRERC.
A possible explanation for this might be that the relation extractor module in CRERC is fine-tuned on 2Wiki; therefore, it can extract entities better than the NER models from Spacy and Flair that are used in our model.
However, the F1 score of our model in the entity-level task is higher than that of CRERC.
\verify{
This indicates that our model can correctly obtain a few triples in a set of gold triples for many samples.}
All our scores (except the F1 score of the entity-level task) are lower than those on NA-Reviewer.
Our target is to analyze the UR tasks in an end-to-end model.
Although the pipeline models (CRERC and NA-Reviewer) are easy to interpret, we cannot determine how the UR tasks affect answer prediction in an end-to-end model.
Therefore, we use the design of our model to perform the analyses in this study.

\subsection{Effectiveness of the UR Tasks}

\paragraph{Reasoning Shortcuts (RQ2)}
\label{r_ques_2}


Table~\ref{reduction_5_time_run} presents the performance drop (smaller is better) for five times running of the four settings of the model on the four debiased sets of 2Wiki and HotpotQA-small. 
As depicted in the table, for 2Wiki, the gap between two settings \#1 (answer prediction task only) and \#4 (all three tasks) is consistent in all five times running.
Meanwhile, for HotpotQA-small, the gap between two settings \#1 (answer prediction task only) and \#4 (all three tasks) is inconsistent in all five times running.
This observation supports our explanation in Section~\ref{analyses_sec} that the position bias in HotpotQA-small does not have a large impact on the main QA task.

\begin{table*}[t]
  \begin{center}
  \resizebox{\textwidth}{!}{%
    \begin{tabular}{l | l | r r | r r | r r | r r | r r } 
    \toprule
      \multirow{3}{3cm}{\textbf{Dataset}} & \multirow{3}{3cm}{\textbf{Task Setting}} &  \multicolumn{10}{c}{\textbf{Reduction (\%)}}  \\ 
      \cmidrule{3-12}
      
     &  & \multicolumn{2}{c}{\textbf{Time \#1}} & \multicolumn{2}{c}{\textbf{Time \#2}} & \multicolumn{2}{c}{\textbf{Time \#3}} & \multicolumn{2}{c}{\textbf{Time \#4}} 
     & \multicolumn{2}{c}{\textbf{Time \#5}}\\
           \cmidrule{3-12}
    &  & EM & F1  & EM & F1 & EM & F1 & EM & F1 & EM & F1 \\
      
     \midrule
   \multicolumn{12}{c}{\textbf{2Wiki}} \\

     \midrule 
    \multirow{4}{3cm}{\textbf{\textsc{AddUnrelated}}} & Ans & \underline{13.26}  &  \underline{12.11}  &  \underline{13.06}  &  \underline{11.87}  &  \underline{13.59}  &  \underline{12.15}  &  \underline{13.31}  &  \underline{12.08}  &  \underline{13.79}  &  \underline{12.44} \\

& Ans + Sent &  10.82  &  9.56  &  10.94  &  9.68  &  10.93  &  9.83  &  11.15  &  9.83  &  11.16  &  9.66 \\

       & Ans + Ent &  \textbf{8.09}  &  \textbf{7.15}  &  \textbf{7.77}  &  \textbf{6.98}  &  \textbf{7.71}  &  \textbf{6.99}  &  \textbf{7.78}  &  \textbf{7.08}  &  \textbf{7.31}  &  \textbf{6.52} \\

& Ans + Sent + Ent &  8.41  &  7.80  &  8.85  &  7.99  &  9.10  &  8.33  &  8.97  &  8.23  &  8.97  &  8.18 \\

     \midrule 
    \multirow{4}{3cm}{\textbf{\textsc{AddRelated}}} & Ans & 3.72  &  3.61  &  3.61  &  3.57  &  3.54  &  3.39  &  3.29  &  3.26  &  3.57  &  3.49 \\

       & Ans + Sent &  \underline{4.22}  &  \underline{4.23}  &  \underline{4.44}  &  \underline{4.44}  &  \underline{4.11}  &  \underline{4.18}  &  \underline{4.08}  &  \underline{4.20}  &  \underline{3.94}  &  \underline{4.04} \\

    & Ans + Ent &  \textbf{2.61}  &  \textbf{2.62}  &  \textbf{2.89}  &  \textbf{2.88}  &  \textbf{2.85}  &  \textbf{2.83}  &  \textbf{2.89}  &  \textbf{2.81}  &  \textbf{2.75}  &  \textbf{2.71} \\

       & Ans + Sent + Ent &  3.18  &  3.12  &  3.18  &  3.14  &  3.06  &  3.06  &  3.23  &  3.25  &  3.14  &  3.07 \\

     \midrule 
    \multirow{4}{3cm}{\textbf{\textsc{Add2}}} & Ans & \underline{12.26}  &  \underline{11.63}  &  \underline{12.59}  &  \underline{12.10}  &  \underline{12.29}  &  \underline{11.69}  &  \underline{12.29}  &  \underline{11.72}  &  \underline{12.16}  &  \underline{11.48} \\

       & Ans + Sent &  11.10  &  10.48  &  11.03  &  10.57  &  11.26  &  10.77  &  11.18  &  10.72  &  11.51  &  10.92 \\

       & Ans + Ent &  \textbf{8.38}  &  \textbf{7.91}  &  \textbf{8.74}  &  \textbf{8.12}  &  \textbf{8.31}  &  \textbf{7.63}  &  \textbf{8.07}  &  \textbf{7.48}  &  \textbf{8.41}  &  \textbf{7.67} \\

       & Ans + Sent + Ent &  9.13  &  8.54  &  8.90  &  8.45  &  8.94  &  8.54  &  8.95  &  8.43  &  9.51  &  8.92 \\

     \midrule 
    \multirow{4}{3cm}{\textbf{\textsc{Add2Swap}}} & Ans & \underline{19.06}  &  \underline{17.61}  &  \underline{18.87}  &  \underline{17.40}  &  \underline{19.20}  &  \underline{17.59}  &  \underline{18.80}  &  \underline{17.31}  &  \underline{19.03}  &  \underline{17.63} \\

       & Ans + Sent &  17.71  &  16.16  &  17.73  &  16.40  &  17.74  &  16.41  &  17.34  &  15.99  &  17.59  &  16.25 \\

       & Ans + Ent &  \textbf{13.02}  &  \textbf{12.19}  &  \textbf{13.09}  &  \textbf{12.19}  &  \textbf{13.38}  &  \textbf{12.30}  &  \textbf{13.13}  &  \textbf{12.30}  &  \textbf{12.97}  &  \textbf{12.07} \\

       & Ans + Sent + Ent &  14.28  &  13.56  &  14.31  &  13.70  &  14.18  &  13.41  &  14.89  &  13.99  &  15.00  &  14.17 \\

     \midrule
   \multicolumn{12}{c}{\textbf{HotpotQA-small}} \\ 

     \midrule 
    \multirow{4}{3cm}{\textbf{\textsc{AddUnrelated}}} & Ans & 4.33 &  \textbf{2.89} &  1.44 &  0.68 &  0.21 &  \textbf{-0.45} &  4.33 &  2.66 &  4.75 &  1.85 \\
    
& Ans + Sent & \textbf{4.01} &  3.65 &  \textbf{0.81} &  \textbf{1.07} &  \textbf{-0.20} &  0.88 &  \textbf{1.01} &  \textbf{0.85} &  \textbf{0.00} &  \textbf{0.28} \\

 & Ans + Ent & 6.17 &  4.97 &  5.97 &  3.98 &  \underline{7.76} &  \underline{7.06} &  \underline{6.38} &  \underline{6.20} &  \underline{7.36} &  \underline{5.80} \\
 
& Ans + Sent + Ent & \underline{6.76} &  \underline{5.83} &  \underline{6.76} &  \underline{5.18} &  1.79 &  2.97 &  4.77 &  4.94 &  5.17 &  4.32 \\

     \midrule 
    \multirow{4}{3cm}{\textbf{\textsc{AddRelated}}} & Ans &   \underline{3.71}  &  \underline{1.14}  &  \underline{4.12}  &  1.10  &  \underline{4.54}  &  \underline{2.12}  &  \underline{3.91}  &  \underline{1.46}  &  \underline{3.91}  &  1.70 \\

  & Ans + Sent & -0.79  &  \textbf{0.10}  &  0.20  &  0.71  &  0.61  &  1.23  &  \textbf{-1.40}  &  \textbf{-0.59}  &  -1.19  &  \textbf{-0.52} \\

  & Ans + Ent &  \textbf{-0.80}  &  0.23  &  \textbf{-0.60}  &  \textbf{-0.19}  &  \textbf{-1.21}  &  \textbf{-0.13}  &  -0.40  &  0.68  &  \textbf{-1.61}  &  -0.42 \\

  & Ans + Sent + Ent & 0.38  &  0.59  &  2.37  &  \underline{1.96}  &  1.19  &  0.95  &  0.00  &  0.59  &  2.37  &  \underline{2.17} \\

     \midrule 
    \multirow{4}{3cm}{\textbf{\textsc{Add2}}} & Ans &  1.04  &  1.01  &  1.04  &  0.17  &  1.04  &  \textbf{0.53}  &  1.64  &  0.51  &  3.50  &  2.83 \\

& Ans + Sent &  1.01  &  1.51  &  \textbf{-0.79}  &  \textbf{-0.19}  &  \textbf{0.00}  &  1.43  &  \textbf{0.00}  &  \textbf{0.32}  &  \textbf{0.20}  &  \textbf{1.20} \\

 & Ans + Ent &  \underline{4.57}  &  \underline{3.59}  &  2.19  &  2.32  &  \underline{5.17}  &  \underline{4.85}  &  \underline{3.78}  &  \underline{3.20}  &  \underline{4.38}  &  \underline{3.73} \\

       & Ans + Sent + Ent & \textbf{0.00}  &  \textbf{0.62}  &  \underline{3.38}  &  \underline{3.63}  &  1.19  &  2.68  &  1.59  &  2.56  &  2.98  &  2.79 \\

     \midrule 
    \multirow{4}{3cm}{\textbf{\textsc{Add2Swap}}} & Ans &  5.16  &  3.55  &  3.10  &  1.40  &  3.71  &  \textbf{1.82}  &  3.29  &  2.12  &  4.54  &  3.45 \\

& Ans + Sent &  3.82  &  3.77  &  \textbf{0.81}  &  \textbf{1.17}  &  2.00  &  2.03  &  \textbf{1.62}  &  \textbf{1.04}  &  \textbf{0.61}  &  \textbf{1.81} \\

   & Ans + Ent &  \underline{5.57}  &  \underline{4.20}  &  \underline{6.38}  &  \underline{5.07}  &  \underline{7.96}  &  \underline{6.83}  &  \underline{6.56}  &  \underline{5.39}  &  \underline{7.96}  &  \underline{5.83} \\

       & Ans + Sent + Ent &  \textbf{3.58}  &  \textbf{3.54}  &  5.77  &  4.74  &  \textbf{1.59}  &  2.52  &  2.37  &  3.04  &  4.57  &  4.38 \\

      \bottomrule
    \end{tabular}
    }
    \caption{
Performance drop (smaller is better) for five times running of the four settings of the model on the four debiased sets of 2Wiki and HotpotQA-small. 
The best and worst scores are boldfaced and underlined, respectively.
The debiased datasets are newly created for each time running.
    }
    \label{reduction_5_time_run}
  \end{center}
\end{table*}


\subsection{Analyses}
\label{analysis_app}

\paragraph{Details of RQ2}
\label{app_position_bias}

\begin{figure}[t]
\centering
    \includegraphics[scale=0.50]{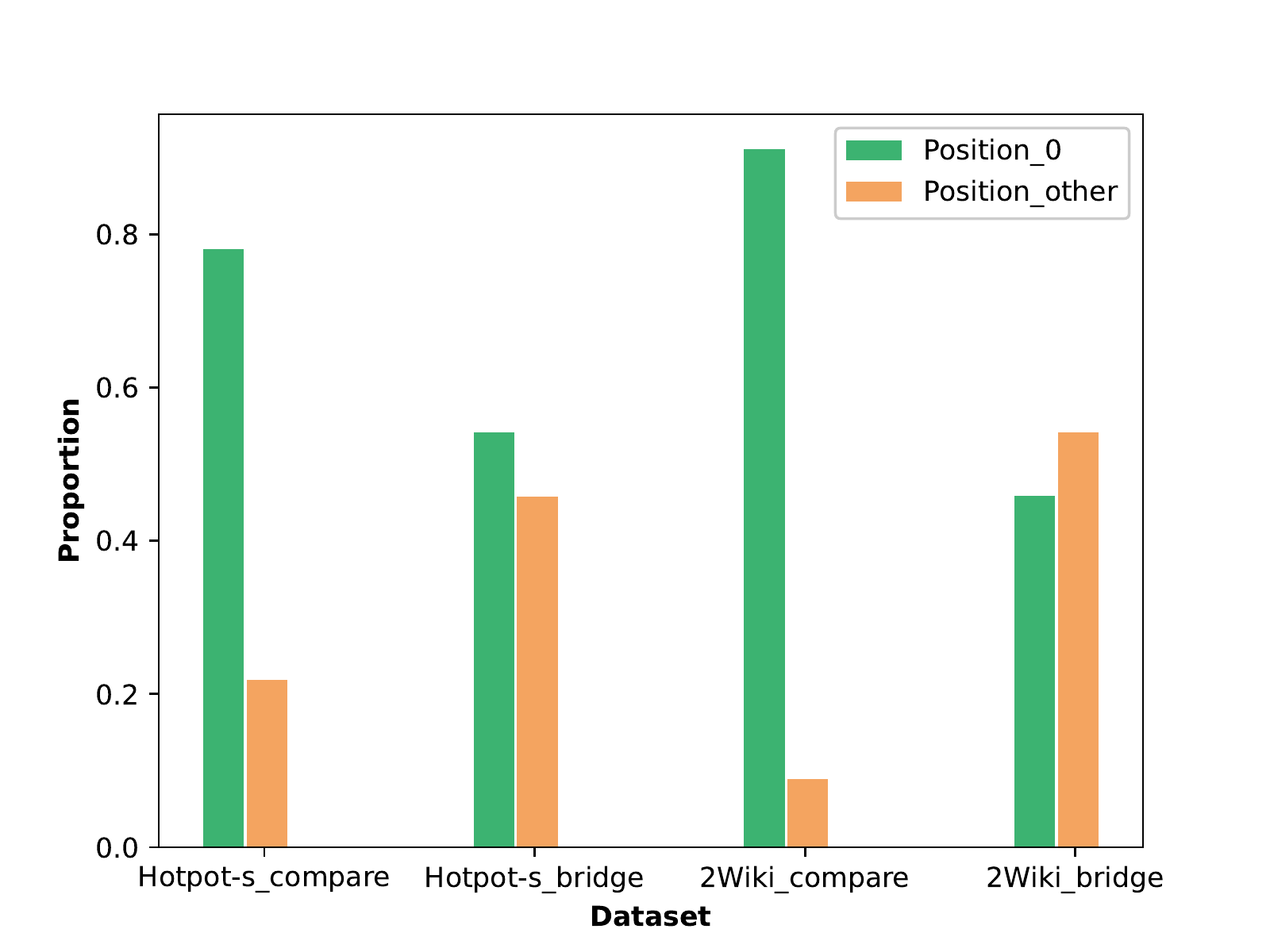}
    \caption{Information on the position of sentence-level SFs of comparison and bridge questions in the dev. sets of the two datasets: 2Wiki and HotpotQA-small.}
    \label{fig:bias_data_ques_type}
\end{figure}

Figure~\ref{fig:bias_data_ques_type} illustrates the information on the position of sentence-level SFs of comparison and bridge questions in the dev. sets of the two datasets: 2Wiki and HotpotQA-small.
As shown in the Figure, the comparison questions have more position biases than the bridge questions in both 2Wiki and HotpotQA-small.
Furthermore, we observe that the position bias in
the comparison questions in HotpotQA-small is smaller than that in 2Wiki.

Table~\ref{reduction_score_ques_type} presents the performance drop for two types of questions, comparison and bridge questions, in 2Wiki and HotpotQA-small.

\begin{table*}[t]
  \begin{center}
  \resizebox{\textwidth}{!}{%
    \begin{tabular}{l | l | r r r r | r r r r } 
    \toprule
      \multirow{3}{3cm}{\textbf{Dataset}} & \multirow{3}{3cm}{\textbf{Task Setting}} &  \multicolumn{4}{c}{\textbf{Comparison}}  & \multicolumn{4}{c}{\textbf{Bridge}}  \\ 
      \cmidrule{3-10}
      
     &  & \multicolumn{2}{c}{\textbf{Answer}} & \multicolumn{2}{c}{\textbf{Answer$\downarrow$} (\%)} & \multicolumn{2}{c}{\textbf{Answer}} & \multicolumn{2}{c}{\textbf{Answer$\downarrow$} (\%)} \\
           \cmidrule{3-10}
    &  & EM & F1  & EM & F1 & EM & F1 & EM & F1 \\
      \midrule

   \multicolumn{10}{c}{\textbf{2Wiki}} \\ 
 \midrule

      \multirow{4}{3cm}{\textbf{Dev}} & Ans &  78.98 &	83.74			  &  &  &  66.10 &	72.85		  &  & \\ 
       & Ans + Sent &  79.45 &	84.21			  &  &  &   67.16 &	73.90			  &  &  \\
       
       & Ans + Ent &  78.86	 & 83.60		   &  &  &  66.75 &	73.61		   &  &   \\
       
       & Ans + Sent + Ent &   80.35	 & 85.08		  &  &  & 	67.84 &	74.49	  &  &    \\      
       
     \midrule
      \multirow{4}{3cm}{\textbf{\textsc{AddUnrelated}}} & Ans &   59.51 &	64.49 &	\underline{24.65} &	\underline{22.99}  &  65.01 &	71.81 &	\textbf{1.65} &	\textbf{1.43} \\ 
       & Ans + Sent & 65.55	 & 71.11 &	17.50 &	15.56 &   64.42 &	71.14 &	\underline{4.08} &	\underline{3.73} \\
       
       & Ans + Ent &  67.67	 & 72.84 &	14.19 &	\textbf{12.87}  &  65.47 &	72.44	& 1.92 &	1.59   \\
       
       & Ans + Sent + Ent &  69.38 &	74.01 &	\textbf{13.65} &	13.01  & 65.72 &	72.48 &	3.13 &	2.70   \\

     \midrule
      \multirow{4}{3cm}{\textbf{\textsc{AddRelated}}} & Ans &  73.60 &	78.22 &	\underline{6.81}	& \underline{6.59} &  65.73 &	72.36 &	\textbf{0.56} &	\textbf{0.67} \\ 
       & Ans + Sent &  74.87 &	79.43 &	5.76 &	5.68  &   65.38	 &71.82 &	\underline{2.65} &	\underline{2.81} \\
       
       & Ans + Ent & 75.57 &	80.17 &	\textbf{4.17} &	\textbf{4.10}  &   66.06 &	72.75 &	1.03 &	1.17   \\
       
       & Ans + Sent + Ent &  76.69 &	81.28 &	4.56 &	4.47  & 	66.63 &	73.14 &	1.78 &	1.81   \\

     \midrule
      \multirow{4}{3cm}{\textbf{\textsc{Add2}}} & Ans & 61.61 &	65.54 &	\underline{21.99} &	\underline{21.73}  &  64.55	 & 71.60 &	2.34 &	1.72 \\ 
       & Ans + Sent & 64.93	 & 69.13 &	18.28 &	17.91  &   64.58 &	71.50 &	\underline{3.84} &	\underline{3.25}  \\
       
       & Ans + Ent & 67.16 &	71.43 &	\textbf{14.84} &	\textbf{14.56}  &   65.51 &	72.52 &	\textbf{1.86} &	\textbf{1.48}   \\
       
       & Ans + Sent + Ent &   67.85	 &72.19 &	15.56 &	15.15  & 66.06 &	72.94 &	2.62 &	2.08  \\

     \midrule
      \multirow{4}{3cm}{\textbf{\textsc{Add2Swap}}} & Ans &   51.13 &	55.50 &	\underline{35.26} &	\underline{33.72}  & 64.42 &	71.55 &	\textbf{2.54} &	\textbf{1.78} \\ 
       & Ans + Sent &  55.19 &	60.21 &	30.53 &	28.50 &  63.96 &	70.83 &	\underline{4.76} &	\underline{4.15} \\
       
       & Ans + Ent & 60.42 &	64.80 &	\textbf{23.38} &	\textbf{22.49}  &   65.04 &	71.99 &	2.56 &	2.20   \\
       
       & Ans + Sent + Ent & 60.25 &	64.37 &	25.02 &	24.34  & 65.51 &	72.23 &	3.43 &	3.03    \\

    \midrule 
   
   \multicolumn{10}{c}{\textbf{HotpotQA-small}} \\

    \midrule
      \multirow{4}{3cm}{\textbf{Dev}} & Ans &  56.44 &	61.86			  &  &  &  51.89 &	67.72		  &  & \\ 
       & Ans + Sent &  57.92 &	63.44			  &  &  &   53.43 &	70.61		  &  &  \\
       
       & Ans + Ent &  57.92 &	63.14		   &  &  &   53.85 &	70.75		   &  &   \\
       
       & Ans + Sent + Ent &  57.43 &	64.44		  &  &  & 	53.99 &	70.86 	  &  &    \\      
       
     \midrule
      \multirow{4}{3cm}{\textbf{\textsc{AddUnrelated}}} & Ans &   50.00 &	56.24 &	11.41 &	9.09  &  50.77 &	66.85 &	\textbf{2.16} &	\textbf{1.28} \\ 
      
       & Ans + Sent &  52.97 &	60.64 &	\textbf{8.55} &	\textbf{4.41}  &   52.03	 &68.17	 & 2.62 &	3.46 \\
       
       & Ans + Ent &  51.49 &	57.43 &	11.10 &	9.04 &   51.33 &	67.97 &	\underline{4.68} &	\underline{3.93}   \\
       
       & Ans + Sent + Ent &   47.03 &	55.59 &	\underline{18.11} &	\underline{13.73}  & 	52.17 &	68.16 &	3.37 &	3.81   \\

     \midrule
      \multirow{4}{3cm}{\textbf{\textsc{AddRelated}}} & Ans &  53.96 &	60.48 &	4.39	& 2.23  &  50.07 &	67.14 &	\underline{3.51} &	\underline{0.86} \\ 
      
       & Ans + Sent & 57.43	& 63.37	 & 0.85	 & 0.11  & 54.13 &	70.54 &	\textbf{-1.31} &	0.10  \\
       
       & Ans + Ent &  58.91 &	64.11 &	\textbf{-1.71} &	\textbf{-1.54}  &   54.13 &	70.27 &	-0.52 &	0.68   \\
       
       & Ans + Sent + Ent &  53.96 &	61.23 &	\underline{6.04} &	\underline{4.98}  & 54.69 &	71.24 &	-1.30 &	\textbf{-0.54}  \\

     \midrule
      \multirow{4}{3cm}{\textbf{\textsc{Add2}}} & Ans &   54.46 &	59.52 &	\underline{3.51} &	\underline{3.78}  & 51.75 &	67.53 &	0.27 &	0.28 \\ 
      
       & Ans + Sent & 58.91	 &64.31	 & \textbf{-1.71}	 & \textbf{-1.37}  &  52.45 &	69.03 &	1.83 &	2.24 \\
       
       & Ans + Ent &  56.93	 &62.33	 &1.71 &	1.28  &   50.91 &	67.81 &	\underline{5.46} &	\underline{4.16}  \\
       
       & Ans + Sent + Ent &   55.94 &	62.58 &	2.59 &	2.89  & 54.41 &	70.82 &	\textbf{-0.78} &	\textbf{0.06}   \\

     \midrule
      \multirow{4}{3cm}{\textbf{\textsc{Add2Swap}}} & Ans &  48.51 &	53.94 &	\underline{14.05} &	\underline{12.80}  &  50.63 &	66.94 &	2.43 &	\textbf{1.15} \\ 
      
       & Ans + Sent &  53.47 &	60.30 &	7.68 &	4.95  &   52.03 &	68.16 &	2.62 &	3.47 \\
       
       & Ans + Ent &  53.96	 &60.51	 &\textbf{6.84} &	\textbf{4.17}  &   51.05	 &67.78 &	\underline{5.20} &	\underline{4.20} \\
       
       & Ans + Sent + Ent &   50.99	 &58.64 &	11.21 &	9.00  & 53.29 &	69.33 &	\textbf{1.30} &	2.16   \\ 
       
      
      \bottomrule
    \end{tabular}
    }
    \caption{
Performance drop (smaller is better) for two types of questions (comparison and bridge questions) of the four settings of the model on the four debiased sets of 2Wiki and HotpotQA-small. 
The best and worst scores are boldfaced and underlined, respectively.
For both 2Wiki and HotpotQA-small, we choose the results from the first time running to perform the analysis.
    }
    \label{reduction_score_ques_type}
  \end{center}
\end{table*}

\paragraph{Details of RQ3}


Table~\ref{example_app} presents examples of the outputs predicted by our model, which is trained on three tasks simultaneously.

\begin{table*}[ht]
\centering
 \resizebox{\textwidth}{!}{%
\begin{tabular}{p{1.7cm} p{14.5cm} }
\toprule

\textbf{Type} & \textbf{Example}  \\ 


\midrule

\multirow{7}{1.7cm}{Bridge - Prune}
&
\multirow{7}{14.5cm}{\textbf{Paragraph A:} {\color{purple} Polish-Russian War} (Wojna polsko-ruska) is a 2009 Polish film directed by Xawery Żuławski based on \ldots  \\
\textbf{Paragraph B:} {\color{blue} Xawery Żuławski} (born 22 December 1971 in Warsaw) is a Polish film director.
\ldots 
He is the son of actress Małgorzata Braunek and director Andrzej Żuławski. \ldots  \\
\textbf{Q:} Who is the director of {\color{purple} Polish-Russian War}? \\
 \textbf{Predicted answer:} Andrzej Żuławski \reduncheck \\
\textbf{Predicted entity-level:} (``Polish-Russian War'', ``director'', ``Xawery Żuławski'') \greencheck 
}
\\  \\ \\ \\ \\ \\  \\

\midrule

\multirow{8}{1.7cm}{Bridge - Prune}
&
\multirow{8}{14.5cm}{\textbf{Paragraph A:} {\color{purple} Francesca von Habsburg} (born 7 June 1958) is an art collector and the estranged wife of Karl von Habsburg, current head of the House of Habsburg- Lorraine.  \\
\textbf{Paragraph B:} {\color{blue} Michaela von Habsburg} was born \ldots  
She is the twin sister of Monika von Habsburg, and daughter of Otto von Habsburg and Princess Regina of Saxe - Meiningen.  \\
\textbf{Q:} Who is the spouse of {\color{purple} Francesca von Habsburg}? \\
 \textbf{Predicted answer:} Princess Regina of Saxe - Meiningen \reduncheck \\
\textbf{Predicted entity-level:} \\ (``Francesca von Habsburg'', ``spouse'', ``Karl von Habsburg'') \greencheck 
}
\\  \\ \\ \\ \\ \\  \\ \\

\midrule

\multirow{9}{1.7cm}{Comparison - Inverted}
&
\multirow{9}{14.5cm}{\textbf{Paragraph A:} {\color{purple} Montréal/Les Cèdres Airport} is a general aviation aerodrome located approximately west of Montreal, Quebec, Canada near Autoroute 20 west of \ldots  \\
\textbf{Paragraph B:} {\color{blue} Flying J Ranch Airport} is a privately owned, public use \ldots 
The airport is located southwest of the central business district of Pima, a city in Graham County, Arizona, United States and northeast of Tucson International Airport. \ldots  \\
\textbf{Q:} Are {\color{purple} Montréal/Les Cèdres Airport} and {\color{blue} Flying J Ranch Airport} located in different countries? \\
 \textbf{Predicted answer:} no \reduncheck \\
\textbf{Predicted entity-level:} (``Flying J Ranch Airport'', ``country'', ``United States'') \& (``Montréal/Les Cèdres Airport'', ``country'', ``Canada'')
\greencheck 
}
\\  \\ \\ \\ \\ \\  \\ \\ \\

\midrule

\multirow{15}{1.7cm}{Comparison - Inverted}
&
\multirow{15}{14.5cm}{\textbf{Paragraph A:} {\color{purple} A Romance of the Air} is a 1918 American silent drama film based \ldots Directed by Harry Revier, the film was \ldots  
\\
\textbf{Paragraph B:} {\color{purple} Harry Revier} (16 March 1890 – 13 August 1957) was ... American director \ldots  \\
\textbf{Paragraph C:} {\color{blue} How Moscha Came Back} is a 1914 silent film comedy short directed by Phillips Smalley. \ldots  \\
\textbf{Paragraph D:} {\color{blue} Phillips Smalley} (August 7, 1865 – May 2, 1939) was an American silent film director and actor.  \\
\textbf{Q:} Which film has the director who was born later, {\color{purple} A Romance of the Air} or {\color{blue} How Moscha Came Back}? \\
 \textbf{Predicted answer:} How Moscha Came Back \reduncheck \\
\textbf{Predicted entity-level:} \\ (``A Romance of the Air'', ``director'', ``Harry Revier''), \\ (``How Moscha Came Back'', ``director'', ``Phillips Smalley''), \\ (``Harry Revier'', ``date of birth'', ``16 March 1890''), \& \\ (``Phillips Smalley'', ``date of birth'', ``August 7, 1865'')
\greencheck 
}
\\  \\ \\ \\ \\ \\  \\ \\ \\ \\ \\ \\ \\ \\ \\

\bottomrule
\end{tabular}
}
\caption{
Examples of the outputs predicted by our model, which is trained on three tasks simultaneously.
}
\label{example_app}
\end{table*}

\end{document}